\documentclass[conference]{IEEEtran}
\usepackage{times}

\pdfoutput=1 
\usepackage{multicol}
\usepackage[bookmarks=true]{hyperref}
\usepackage{graphicx}
\usepackage{multirow}
\usepackage{adjustbox}
\usepackage{multirow}
\usepackage{tabularx}
\usepackage{pdflscape}
\usepackage{booktabs} 
\usepackage{xcolor}

\begin{document}

\title{Egocentric Robots in a Human-Centric World? Exploring Group-Robot-Interaction in Public Spaces}

\author{Ana Müller and Anja Richert}

\author{\authorblockN{Ana Müller}
\authorblockA{Cologne Cobots Lab\\
TH Köln - University of Applied Sciences\\
Cologne, Germany\\
ana.mueller@th-koeln.de}
\and
\authorblockN{Anja Richert}
\authorblockA{Cologne Cobots Lab\\
TH Köln - University of Applied Sciences\\
Cologne, Germany\\
anja.richert@th-koeln.de}}

\maketitle

\begin{abstract}
The deployment of social robots in real-world scenarios is increasing, supporting humans in various contexts. However, they still struggle to grasp social dynamics, especially in public spaces, sometimes resulting in violations of social norms, such as interrupting human conversations. This behavior, originating from a limited processing of social norms, might be perceived as robot-centered. Understanding social dynamics, particularly in group-robot-interactions (GRI), underscores the need for further research and development in human-robot-interaction (HRI). Enhancing the interaction abilities of social robots, especially in GRIs, can improve their effectiveness in real-world applications on a micro-level, as group interactions lead to increased motivation and comfort. In this study, we assessed the influence of the interaction condition (dyadic vs. tridaic) on the perceived extraversion (ext.) of social robots in public spaces. The research involved \(40\) HRIs, including \(24\) dyadic (i.e., one human and one robot) interactions and \(16\) triadic interactions, which involve at least three entities, including the robot.

\end{abstract}

\IEEEpeerreviewmaketitle

\section{Motivation and Related Work}

The use of social robots in real-world scenarios is growing, as they are designed to engage and foster social connections with humans. While social robots are already effectively in immediate initiating verbal and non-verbal interactions upon recognizing a human presence (i.e., a potential user), their complete acceptance remains pending. A contributing factor might be their insufficient ability to comprehend the intricacies of human behavior and the demands of social interactions, shaped by the social norms and conventions prevalent in interactions between humans [3]. Certain systems, such as robots that navigate in public spaces, can identify groups of people and the interactions between individuals, considering social norms: Some resort to the ‘\emph{facing-formation}’ [14], shaped by social norms, and navigate not through, but around, a group of people [20, 4, 27]. Others can recognize groups of people in open environments from a robot-centric perspective (i.e., through their camera) using unsupervised learning to determine how individuals and groups are likely to move [26, 15]. Recognizing social norms is important at this point, but this is frequently just a practical consideration, such as preventing collisions to ensure that social robots are sufficiently robust and safe for autonomous operation in open environments [3]. In addition to navigation skills, the consensus in the field of human-robot-interaction (HRI) regarding the need to carry out studies in unpredictable environments of public spaces to improve the social behavior of such systems - static or mobile [11] - is growing, as it offers greater views on varied user demands and intricate social relationships [22, 18, 27]

\begin{figure}[htbp]
    \centering
    \includegraphics[width=1.0\linewidth]{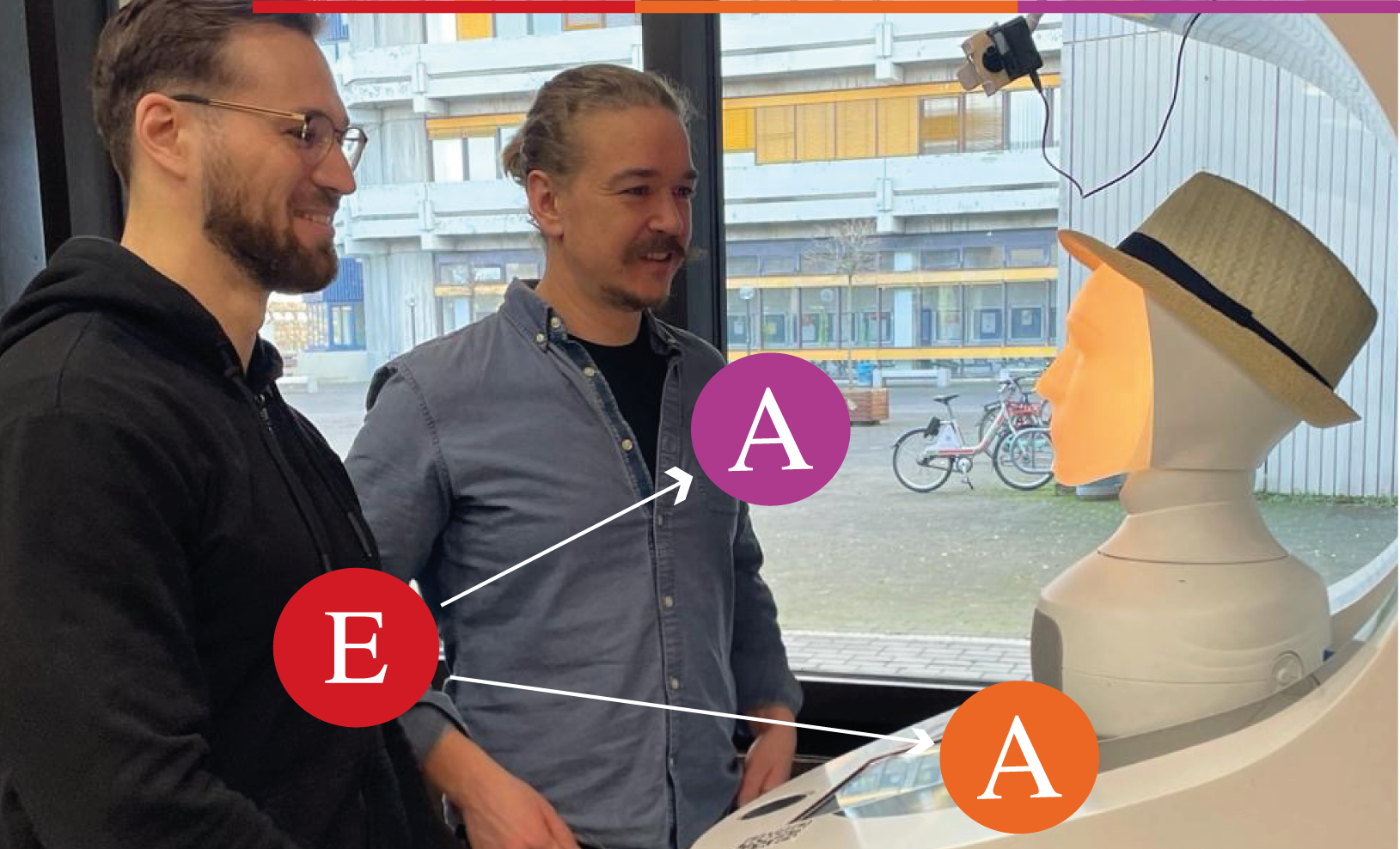}
    \caption{Interaction with Furhat at TH Köln. Showcasing research via ENA, involving one participant as the ego (survey respondent) and others, including the robot, as alteri. The interaction is described from the ego's viewpoint. The individuals depicted are not actual study participants to maintain privacy.}
    \label{Inteaction}
\end{figure}

The insights gained can be used to derive success factors for the use of social robots in real-world scenarios, to enable robots to adapt to human social norms and provide a consistent set of behaviors [3, 27]. However, strategies for implementing these social norms as design characteristics are limited [5, 27], and significant challenges remain, such as understanding the distinct social dynamics in group-robot-interactions (GRI) [11], as previous work has indicated that HRI often occurs collaboratively [9, 21]. This is important, as dyadic interactions (i.e., one human and one robot) tend to be robust, but GRIs present different challenges due to their inherent social intricacies [16]. Nevertheless, the HRI-community has focused mainly on dyadics [18, 1, 11], slightly overlooking improvements in communication and interaction in GRI. However, in GRI, also the demands for acknowledging social norms and conventions are significantly higher and present a greater challenge for social robots to handle. For example, social robots are currently unable to discern when individuals are conversing within groups and presently show little enthusiasm for engaging with the robot. This gap in recognizing the dynamics at play, currently contributes to the fact that robots sometimes violate social norms; e.g., they (continue to) speak while humans are talking (to each other), and most of them have no way to react dynamically, e.g., to speaker changes. Along with this, there has been calls for research on how to improve turn-taking in conversational systems (not limited to social robots) [25] for dyadic and also for group interactions. For example, Żarkowski [29] investigated approaches for compliance with human turn-taking norms and discovered that this led to a decrease in conversational mishaps, thus improving communication effectiveness in GRI. Furthermore, the area of adaptive conversation designs has seen a recent increase in interest, driven by developments in generative AI to specifically support GRI (e.g., [2, 19]. However, these studies were conducted under laboratory conditions without the dynamics at play in real-world scenarios. 

As humans are very good at coordination and achieve fluent turn-taking [25], one might conclude that the design of social robots and the lack of understanding of social conventions can be regarded as egocentric behavior. The term ‘egocentrism’ traditionally refers to a human trait, but in this context, it aptly describes social robots that are currently blind to social complexities and disguise this blindness through actionism, i.e., constant attempts to establish relationships with humans. This behavior reflects a robot-centric perspective in interaction design, not similar to the robot-centric approach in navigation, as described by Taylor et al. [26], who consider robot centricity as a design feature in the successful robotics navigation capability. As individuals are more often fully involved in the HRI, this can be a greater influencing factor in GRI. This underscores the need for further research and development (R\&D) to enhance their understanding of social dynamics in unpredictable environment at the macro-level of GRI. 

In this study, we assessed the influence of the interaction condition (dyadic vs. tridaic) on the perceived extraversion (ext.) of social robots in public spaces. In the university's cafeteria the research examined \(40\) interactions with a social robot, consisting of \(24\) dyadic and \(16\) triadic (i.e., GRI). The research design acknowledges the realities of field experiments, recognizing that not all users are willing to participate in surveys. Therefore, we assessed GRI using ego-network analysis (ENA). Originating from empirical social sciences, ENA collects comprehensive data on topics such as voting patterns in elections. Generally, respondents in these studies (ego) reveal their voting intentions. Following this, a prompt, i.e., name generator [28], is used to prompt respondents to identify specific individuals from their networks (alteri) and subsequently evaluate the voting behavior of these individuals as well [10]. In our research approach, respondents (egos) complete a detailed self-evaluation of their interaction with the robot. They also assess the alteri (i.e., group members) and the robots behavior in both group and dyadic settings. This allows us to see a network of connections between the participant, the robot and – if apparent – other group members, even if not all group members fill out the questionnaire. The primary goal was to improve our understanding of social dynamics and the impact of group structures on H/GRI. Enhancing this understanding can also aid in the overall advancement of HRI in practical settings, as individuals in GRIs are more likely to be motivated to interact [8], exhibit higher engagement levels, and feel more at ease [24]. As a result, effective GRI will further support the advancement of dyadic HRI at the micro-level.

\section{Materials and Methods} \label{methods}

The study was conducted at the Deutz campus of TH Köln - University of Applied Sciences, Germany, using a Furhat robot (Furhat Robotics) to examine H/GRI in the university's cafeteria on \(13\) experimental days in Jan. and Feb. 2024, during the cafeteria's opening hours. The system involved a Furhat placed on a stand with a screen to display transcribed dialogues (Fig. 1).

\begin{figure}[htbp]
    \centering
    \includegraphics[width=1.0\linewidth]{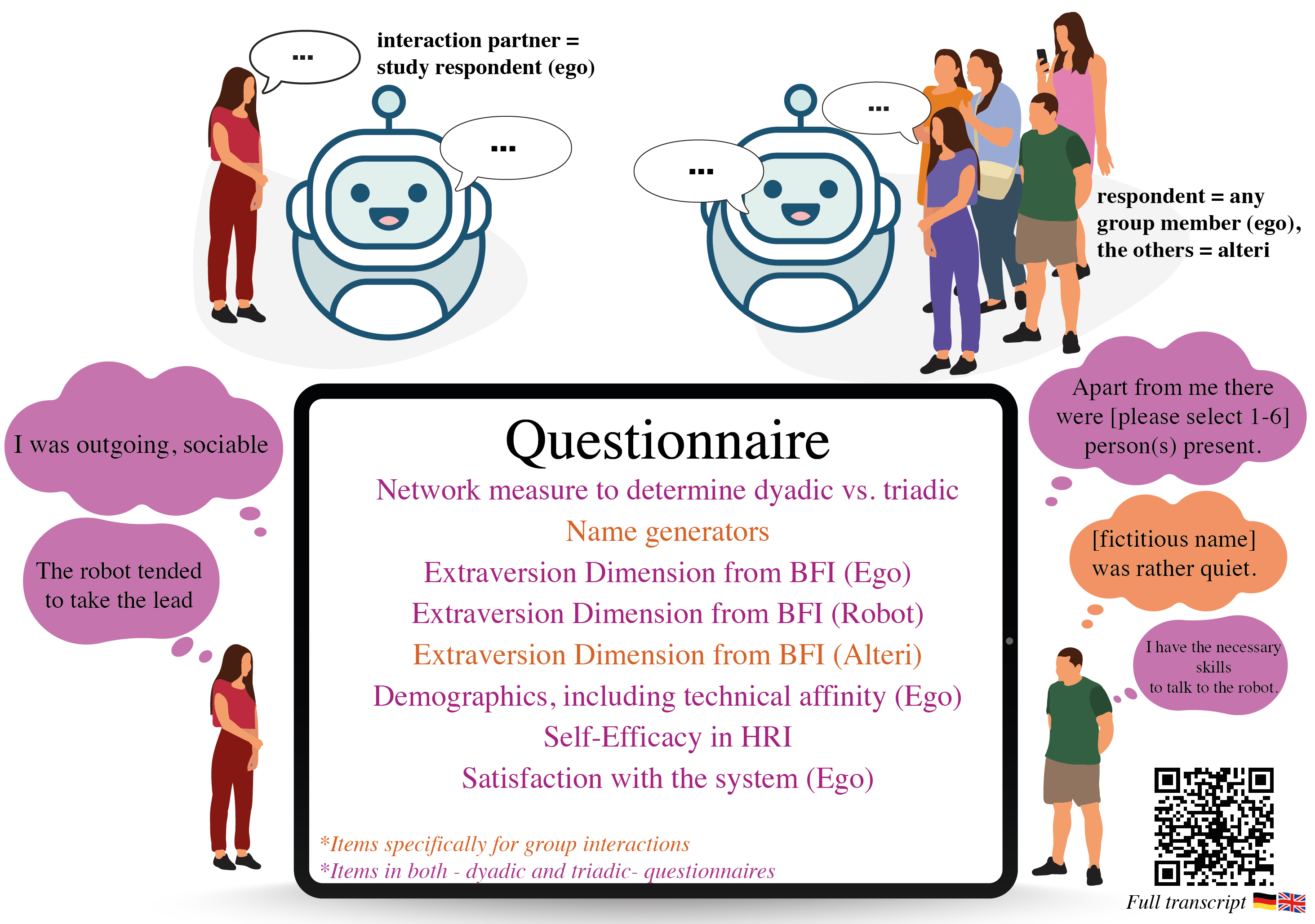}
    \caption{Research methodology using ENA to assess GRIs. Top part: interaction conditions. Bottom part: Questionnaire categories relevant to all interaction conditions (purple) and those specific to GRI (orange). The questionnaire transcript is available for download by scanning the QR code.}
    \label{ResearchDesign}
\end{figure}

The robot was connected to a self-developed artificial intelligence (AI)-powered dialogue system that allowed for interactions in German and English in multi-turn dialougues. YOLOv5m [12] collaborated with Furhat’s person detection to differentiate between dyadic and GRI, addressing this distinction in the conversation design, including the distinction between ‘you’ singular (‘du’) and ‘you’ plural (‘ihr/euch’) in German. The system operated on a curated knowledge base that included information on navigation, cafeteria offerings, robotics, and small talk. Generative AI (GPT 3.5, OpenAI) was used to refine training data sets and develop a conversation design for GRI. Additionally, Llama-2 (Meta) was integrated as a secondary resource for requests beyond the knowledge base's scope. The R\&D received approval from the Research Ethics Committee of TH Köln (application no. THK-2023-0004). The research design, illustrated in Fig. 2, acknowledges the realities of field experiments, using ENA to capture relational dynamics in GRI without requiring every group member to complete a questionnaire. Our methodology involves egos (i.e., questionnaire respondents) providing a comprehensive self-assessment of their interaction and evaluations of the alteri, including the robot (Fig. 1). 

The research included \(40\) randomly chosen participants who interacted with the robot freely and filled out a questionnaire at the outset. We analysed the structure and network density to distinguish between dyadic and GRI (i.e., respondents initially reported the count of individuals interacting with the robot alongside them,  Fig. 2) and implemented a verification step to allow corrections to the reported group sizes if needed. Groups were defined based on a sociological definition, which considers any systematic pattern of interaction involving at least two individuals as a group [23, 7]. Hence, GRI-networks are defined as at least two humans interacting with the robot under a shared session ID. The data collected encompassed details from \(24\) dyadic interactions and \(16\) GRIs. GRI-respondents assigned fictitious names to the (human) alteri for consistent identification throughout the questionnaire. The study utilized the extraversion (ext.) scale from the \emph{Big Five Inventory (BFI-2)} (N of Items = \(12\)) , German Edition [6], a personality trait framework that describes individuals high in extraversion as outgoing, sociable, and vibrant [13], and incorporates elements related to personal as well as group behavior (e.g., “I let the others make the decisions”) to understand the structure of interactions with the robot. In the context of GRI, highly extraverted entities (i.e., humans as well as the robot) might be more active, communicative, and central in interactions, whereas those with lower ext. might be more subdued and less likely to seek the spotlight. The questionnaire was modified to reflect the interaction scenarios with the robot. Respondents evaluated their own ext. (e.g., “During the interaction... I tended to be quiet”), their view of the robot's ext. (e.g., “... the robot tended to take the lead”), and the ext. of each alteri (e.g., “... [fictive name] was rather shy”) using a 5-point Likert scale, with higher scores indicating stronger agreement. After adjusting for item polarity, the reliability analysis on self-reported ext. yielded a Cronbach's $\alpha$ of \(.761\), indicating acceptable reliability. Similarly, the evaluation of the robot's ext. by the egos confirmed internal consistency with a Cronbach's $\alpha$ of \(.703\). The ext. ratings of the alteri demonstrated a notably high Cronbach's $\alpha$ of \(.950\), which will be discussed in further in Sec. III. Additionally, ego provided demographic data (age, education, occupation) and details on \emph{Self-Efficacy} concerning required skills in using the robot and the comparison of their communication ease with others (final item scored inversely) using Ninomiya et al. [17] on a 5-point Likert scale, where a higher score reflects greater agreement. Egos also reported on their use of intelligent personal assistants (IPA) (e.g., Alexa) and generative AI (e.g., ChatGPT) using the same 5-point Likert scale. Further, a scale was developed to measure the openness versus reluctance towards AI, along with their experiences with these technologies, using sliders ranging from \(0-100\%\), showing high internal consistency (Cronbach's $\alpha$ \(= .836\)). Additionally, a scale was created to evaluate the ego's experiences with the robot, examining eight attributes on a 5-point Likert scale, where a higher score indicated stronger agreement: system competence, helpfulness, empathy, human-likeness, likeability, trustworthiness, linguistic competence, and autonomy. This resulted in a Cronbach's $\alpha$ of \(.852\) (\(21\) Items), confirming internal consistency validating the scale's effectiveness in gathering user satisfaction with the robot.

\section{Results and Discussion}\label{results}

The data set consisted of \(24\) dyadic and \(16\) GRI, observed from the egos' viewpoint. Each GRI had between two and three members, totaling \(57\) people (i.e., egos + alteris) who interacted with the robot and were included using ENA. The average age of the egos was \(27.84\) years. Among these, \(35\) were students, one of them part-time and six were full-time employees. The educational background was homogeneous: \(57.5\%\) had a university entrance qualification and \(42.5\%\) had at least one university degree. 

The details of descriptive analysis are shown in Tab. I. Concerning the use of IPAs, the egos showed various reactions. Compared diversity was observed in the use of generative AI. All egos were familiar with these technologies and had used any of these at least once. However, analysis indicates that egos in dyadic interactions used IPAs more frequently than those in GRIs. In contrast, the frequency of generative AI usage and receptiveness to it were higher among those who interact with the Furhat in GRI. The \emph{Self-Efficacy} scale for robot usage indicated a consistent perception, with marginally lower self-efficacy scores for the ego in GRI, suggesting a comparison with the abilities of other group members. Descriptive statistics revealed different levels of satisfaction among users with the robot's performance. This analysis showed that the satisfaction values for the GRIs were slightly lower than for dyadic interactions. This distinction suggests nuanced differences in user satisfaction in dyadic versus GRIs and points to the complexity of users' expectations and experiences when interacting with the system in different conditions.

\begin{table*}[htbp]
\centering
\caption{Descriptive Data Sorted by Interaction Condition}
\label{deskript}
\begin{tabularx}{\textwidth}{@{}l>{\centering\arraybackslash}X>{\centering\arraybackslash}X>{\centering\arraybackslash}X@{}}
\toprule
\textbf{Items} & \textbf{Dyadic \((n=24)\)} & \textbf{Group \((n=16)\)} & \textbf{Total \((N=40)\)} \\
\midrule
Gender (f/m/d) & \(9/15/-\) & \(7/7/1\) (\(1\) unrep.) & \(16/22/1\) (\(1\) unrep.) \\
Age  & \(29.09\) \((11.94)\), (\(2\) unrep.) & \(26\) (\(3.89\)), (\(1\) unrep.) & \(27.84\) (\(9.56\)), (\(3\) unrep.) \\
Use of IPAs (M) & \(3.46\) & \(3.19\) & \(3.35\) \\
Use of gen. AI (M) & \(3.63\) & \(4.07\) & \(3.79\) \\
Attitudes towards AI (\% M (SD)) & \(69.88\) (\(30.29\)) & \(76.75\) (\(24.33\)) & \(72.63\) (\(27.93\)) \\
System satisfaction scale (M (SD)) & \(3.49\) (\(.58\)) & \(3.3\) (\(.50\)) & \(3.41\) (\(.55\)) \\
Self-Efficacy scale (M (SD)) & \(3.99\) (\(.72\)) & \(3.78\) (\(.75\)) & \(3.90\) (\(.73\)) \\
\midrule
Ext. of ego (M (SD)) & \(3.18\) (\(.68\)) & \(3.22\) (\(.83\)) & \(3.20\) (\(.73\)) \\
Ext. of the robot (M (SD)) & \(3.16\) (\(.52\)) & \(2.84\) (\(.61\)) & \(3.03\) (\(.58\)) \\
Ext. of the alteri (M (SD)) & - & \(2.89\) (\(1.09\)), (\(2\) unrep.) & - \\
\bottomrule
\end{tabularx}
\end{table*}

As described in Sec. II, we assessed the perceived levels of extraversion (ext.) among respondents (egos), robot-alteri and human-alteri (if applicable) within different interaction conditions. Highly extraverted entities might be more active, communicative, and central in interactions, whereas those with lower ext. might be more subdued and less likely to seek the spotlight. The descriptive statistics (Tab. I) reveal moderate to slightly positive assessments of ext. across all entities involved, with subtle variations between the interaction conditions. In GRIs, the ext. scores for alteris varied widely, underscoring the diversity of individual perceptions of ext., with a range from \(1.10\) to \(4.90\). Although the sample was structured homogeneously, a more homogeneous perception of alteris’ ext. might imply potential redundancies in the items, given the high internal consistency of the alteris’ ext. scale, as indicated in Sec. II. For egos, there was a consistent perception of moderate ext., with a slight increase in GRIs compared to dyadic interactions. This might suggest a tendency for egos to perceive themselves as slightly more extraverted or to exhibit more extraverted behavior in the presence of other humans. Regarding the robot’s ext., it was consistently perceived as moderately extraverted, with a slight decrease in GRI compared to dyadic interactions. This decrease might reflect varied perceptions of the robot’s behavior in different social settings.

\begin{table}[htbp]
    \centering
    \caption{Perceptions of Self-Extraversion and Robot's Extraversion Across Interaction Conditions}
    \begin{tabular}{lcccc}
        \toprule
        & \multicolumn{2}{c}{Self-Extraversion} & \multicolumn{2}{c}{Robot's Extraversion} \\
        \cmidrule(lr){2-3} \cmidrule(lr){4-5}
        Condition & Mean $\pm$ SD & N & Mean $\pm$ SD & N \\
        \midrule
        GRI (N of Items \(12\)) & \(3.22\) $\pm$ \(.83\) & \(16\) & \(2.84\) $\pm$ \(.61\) & \(16\) \\
        Dyadic (N of Items \(12\)) & \(3.19\) $\pm$ \(.68\) & \(24\) & \(3.16\) $\pm$ \(.53\) & \(24\) \\
        \midrule
        \multicolumn{5}{l}{\(t(38) = .152\), \(p = .88\) for self-extraversion} \\
        \multicolumn{5}{l}{\(t(38) = -1.764\), \(p = .086\) for robot's extraversion} \\
        \bottomrule
    \end{tabular}
    \label{Tab}
\end{table}

To assess the effects of interaction conditions on how egos view their own ext. and their perception of the robot’s ext., we further conducted independent sample t-tests, supported by initial checks for normality and equal variance in dyadic and GRI. Our results (Tab. II) did not show significant differences in the average ext. scores self-assessed by egos between dyadic and GRI, with \(t(38) = .152\), \(p=.88\). The uniformity in self-perceived ext. across HRI conditions indicates a consistent self-perception in varying interactive settings. As we hypothesized that the gap in recognizing the dynamics at play in GRI and the robots’ violation of social norms can be interpreted as egocentric behavior, we similarly analyzed the perception of robots’ ext. depending on the HRI condition with independent sample t-tests, again supported by initial checks for normality and equal variance. However, similarly, examination of the robot’s perceived ext. revealed no significant differences between these conditions, as shown by a t-test result of \(t(38) = -1.764\), \(p=.086\) (Tab. II). These findings suggest a stable level of perceived ext. of the robot in different HRI conditions, highlighting that neither individuals’ self-perception of ext. nor their perception of the robot’s ext. significantly changes in triadic versus dyadic settings.

\section{Conclusion} \label{conclusion}

Our study examined extraversion (ext.) perceptions among respondents (egos), robot-alteri, and human-alteri in different interaction conditions, based on responses from \(40\) egos engaged in dyadic and GRIs with a Furhat robot in the semi-public space of the university. The insights gleaned from the descriptive statistics reveal a nuanced picture of how individuals perceive their own ext. and that of robots in varying within in-the-wild HRI conditions, such as groups of different sizes. Parametric statistical tests showed that the egos exhibited a consistently moderate assessment of self- and robot ext., with no significant differences observed in dyadic versus GRI. In particular, the variation in responses to the alteris ext. suggests individual differences in how group dynamics are perceived and navigated, reflecting the relevance of the scale using ENA to assess. Moreover, the importance of effective turn-taking and dynamic interaction capabilities in conversational systems, as highlighted by e.g., Skantze [25] and Żarkowski [29], is evident in our results. The consistent perception of moderate extraversion of the robot in different conditions suggests that while the robot can engage in interactions, its behavior may not yet be sufficiently adaptive to different social settings, particularly in group conditions, resulting in slightly lower system satisfaction. This underscores the need for advancements in adaptive conversation designs and the integration of generative AI to support GRIs, building on e.g., Addlesee et al. [2] and Paetzel-Prüsmann and Kennedy [19]. 

While our study strikes through its in-the-wild design, it encounters several limitations, such as the small sample size and the difference in the sample sizes between the dyadic and GRI samples could affect the statistical power of the t-test, potentially limiting its ability to detect small but meaningful differences between the conditions. However, limitations point out the opportunity for future studies to build on our initial findings, preferably with larger and more heterogeneous samples and data collection throughout the alteris. However, our results support further exploration of personality dynamics within HRI, to support the development of social robots capable of navigating the complexities of human social environments in the near future.

\section*{Acknowledgments}

The authors acknowledge the financial support by the Federal Ministry of Education and Research of Germany in the framework FH-Kooperativ 2-2019 (project number 13FH504KX9). We thank our collaboration partner DB Systel GmbH and all other collaborators for their contributions.

\end{document}